\def\year{2021}\relax
\documentclass[letterpaper]{article} 
\usepackage{aaai21}  
\usepackage{times}  
\usepackage{helvet} 
\usepackage{courier}  
\usepackage[hyphens]{url}  
\usepackage{graphicx} 
\def\UrlFont{\rm}  
\usepackage{natbib}  
\usepackage{caption} 
\usepackage{times}  
\usepackage{helvet} 
\usepackage{courier}  
\usepackage[hyphens]{url}  
\usepackage{graphicx} 
\def\UrlFont{\rm}  
\usepackage{latexsym}

\usepackage{times}
\usepackage{soul}
\usepackage{url}
\usepackage[utf8]{inputenc}
\usepackage{amsmath}
\usepackage{booktabs}
\usepackage{algorithm}
\usepackage{algorithmic}
\usepackage{url,ifthen}
\usepackage{amssymb}

\usepackage{fancybox}
\usepackage{tikz}

\usepackage{grbbridge}
\usepackage{enumitem}

\usepackage{color}

\usepackage[switch]{lineno} 
\begin{document}

\title{Optimizing $\alpha\mu$}

\author{Tristan Cazenave\textsuperscript{\rm 1},
        Swann Legras\textsuperscript{\rm 2},
        Véronique Ventos\textsuperscript{\rm 2}\\
}

\affiliations{
    \textsuperscript{\rm 1} LAMSADE, Université Paris-Dauphine, PSL, CNRS, France\\ \textsuperscript{\rm 2} Nukkai, France\\
    Tristan.Cazenave@dauphine.psl.eu
    slegras@nukk.ai 
    vventos@nukk.ai
}

\maketitle

\begin{abstract}
$\alpha\mu$ is a search algorithm which repairs two defaults of Perfect Information Monte Carlo search: strategy fusion and non locality. In this paper we optimize $\alpha\mu$ for the game of Bridge, avoiding useless computations. The proposed optimizations are general and apply to other  imperfect information turn-based games. We define multiple optimizations involving Pareto fronts, and show that these optimizations speed up the search. Some of these optimizations are cuts that stop the search at a node, while others keep track of which possible worlds have become redundant, avoiding unnecessary, costly evaluations. We also measure the benefits of parallelizing the double dummy searches at the leaves of the $\alpha\mu$ search tree.
\end{abstract}

\section{Introduction}

The state of the art for imperfect information card games is Perfect Information Monte Carlo sampling (PIMC). It was first proposed by Levy \cite{levy1989million} for Bridge, and used in the popular program GIB \cite{ginsberg2001}. PIMC can be used in other trick-taking card games such as Skat \cite{buro2009improving,kupferschmid2006skat}, Spades and Hearts \cite{sturtevant2006feature}. Long analyzed the reasons why PIMC is successful in these games \cite{long2010understanding}. The principle of PIMC is to use determinization to generate possible worlds. For each possible move,  PIMC plays the move, samples from a set of possible worlds, and then solves each world exactly as if it was a perfect information game. It then computes the average of the results among the possible worlds to evaluate each move.

Other approaches to imperfect information games are Information Set Monte Carlo Tree Search \cite{cowling2012information}, counterfactual regret minimization that solved Poker \cite{zinkevich2008regret,brown2019superhuman}, and Exploitability Descent \cite{lockhart2019computing}. There are also various Reinforcement Learning algorithms for incomplete information games are available in the OpenSpiel framework \cite{lanctot2019openspiel}.

Recursive Monte Carlo Search \cite{furtak2013} is the adaptation of Nested Monte Carlo Search \cite{CazenaveIJCAI09} to multi-player and incomplete information games. The basic principle is the same: it uses Monte Carlo Search to improve Monte Carlo Search made at a higher level of recursion. Recursive Monte Carlo Search improves on PIMC for Skat and Bridge \cite{bouzy2020}, however, it takes much more time to complete than PIMC.

PIMC plays sub-optimally due to two main problems: strategy fusion and non-locality \cite{frank2001theoretical}. $\alpha\mu$ \cite{cazenave2020alphamu} is an anytime heuristic search algorithm for incomplete information games that assumes perfect information for the opponents. $\alpha\mu$ addresses the strategy fusion and non-locality problems encountered by PIMC. Other programs also address the strategy fusion problem in endgame play, for example, GIB \cite{ginsberg2001} in Bridge  (using a single dummy solver) and the Skat endgame solver of Stefan Edelkamp \cite{edelkamp2020representing}.

In this paper we propose improvements to speedup the $\alpha\mu$ algorithm without affecting it's output.

The paper starts with a section on previous work in Bridge and on the $\alpha\mu$ algorithm. The next section describes the optimizations and the modified algorithm. The last section gives experimental results.

\section{Prerequisites and Previous Work}

In this section, we introduce some definitions used in the paper, we then briefly explain the game of Bridge, Computer Bridge, previous work on dealing with strategy fusion, non-locality, the early cut and the root cut that were defined previously in the $\alpha\mu$ paper \cite{cazenave2020alphamu}.

\subsection{Definitions}

We define general naming conventions used throughout the article.

\subsubsection{Vectors}

Given $n$ different possible worlds, a boolean vector of size $n$ keeps the status of the game for each possible world: A zero at index $w$ means that the game is lost for world number $w$. A one means the game is won. Associated to the vector there is another vector of booleans indicating whether the world is possible in the current state. At the root of the search all worlds are possible but when an opponent makes a move, the move is usually only valid in some of the worlds and the set of valid worlds is reduced.

\subsubsection{Pareto Fronts}

A set of vectors that are not currently dominated by other sets of vectors.





We say a Pareto front $f_1$ dominate $f_2$ iff:\\


$\forall x_2 \in f_2, \exists x_1 \in f_1, x_1$ dominates $x_2$ or $x_1 = x_2$.\\


We say a vector $x_1 \in \{0,1\}^n$ dominates $x_2$ iff:\\

$\forall i  \in [1,n]$: $x_1[i] \geq x_2[i]$ and\\ 

$\exists i \in [1,n]$ such that $x_1[i] > x_2[i].$\\

The score of a vector is the average among all possible worlds of the values contained in the vector.

\subsubsection{Impossible worlds}

When the search reaches a node where some worlds have disappeared because the moves that have been played are not possible in these worlds. These worlds are represented as ''x'' in figures in this article. New impossible worlds appear at min nodes when we consider a move from the defender possible in only a subset of all worlds.

\subsubsection{Useless worlds}

Worlds for which we know for sure that nothing is lost by labelling them 0. These worlds are noted ''-'' on figures. 

\subsubsection{Comparison of Pareto Fronts}

When comparing two Pareto fronts from different depths in the tree or at min node between two actions that contain different impossible worlds, to test for the dominance of one over the other, one must carefully handle impossible worlds as they can be a win higher in the tree. Such impossible worlds must be evaluated as 1 unless proven useless higher in the tree when checking for dominance relation. As such, [1~x~0] is not dominated by [1~0~0] but is dominated by [1~1~0] or even by [1~x~-].

\subsection{Bridge in short}

The interested reader can refer for instance to \cite{mahmood2014bridge} for a more complete presentation of the game of Bridge.\\ 
Bridge is a trick-taking card game with four players (denoted by West, North, East  and South or W,N,E,S)  divided in two partnerships (East-West and  North-South).  A standard 52 card deck is shuffled and each player receives a hand of 13 cards that is only visible to them. A Bridge deal  (or \textit{board}) is divided into two major phases: the bidding (out of the scope of the paper) and the card play.

The goal of the bidding is to reach a contract which determines the minimum number of tricks the pair commits to win during the card play, either with no trump (NT) or with a determined suit as trump.

In the following, we assume that the North-South pair reached the contract of 3NT (resp. 7NT) and that South is the agent who plays the board (i.e the \textit{declarer}).  

During the card play, the goal is to fulfill (for the declarer) or to defeat (for the \textit{defenders}) the contract reached during the bidding phase. For the contract of 3NT (resp. 7NT),  the minimum number of tricks required to win the board is 9 (resp. 13).

The player on the left of the declarer exposes the first card  of the game. The declarer's partner (called the \textit{Dummy}) then lays their cards face up on the table. 

When playing in a NT contract, there is only one simple rule : each player is required to follow suit if possible and  can play any card of the suit.

When the four players have played a card, the player who played the highest-ranked card in the suit  (2$<$3$<$...$<$10,J,Q,K,A)  wins the trick and they will  be on lead at the following trick.

The board is over when all the  cards have been played.

\subsection{Computer Bridge}

Since 1996, the best bridge programs can annually participate in the World Computer-Bridge Championship (WCBC). The selected bridge AI used in our experiments was developed by Yves Costel for the bridge program Wbridge5\footnote{\url{http://www.wbridge5.com}}.
The boosted version of Wbridge5 \cite{ictai2017} won the WCBC in 2016, 2017 and 2018. 

The best Bridge bots use PIMC, with a double-dummy solver (DDS) to evaluate each simulated world.  They also use some additional heuristics, for example to prefer the less-revealing of equivalent actions.

DDS  gives the number of tricks won by each side for Spade, Heart, Diamond, Club or No Trump contracts  when all four players know the placement of the 52 cards, and each player plays optimally. 

A tree-search can then be used in this simplified game, as well as standard alpha-beta pruning methods. Values of the leaves are computed using the double-dummy solver. The growing number of tricks won during the play is used as upper and lower bounds in the alpha-beta process. 

A very efficient DDS has been written (and made public) by Bo Haglund \cite{haglund2010search}.

Currently, the  average level of the best current Bridge AIs are still far from the level of professional players, and closer to the level of good amateurs. This is partly due to the aforementioned problems of PIMC: namely strategy fusion and non-locality.

The strategy fusion problem of PIMC comes from the fact that DDS plays different Max moves in the different worlds of an information set. In the real game, Max has to play the same move in all the worlds of an information set. $\alpha\mu$ addresses this problem by evaluating every Max move jointly for all the possible worlds in the information set during its search. 

The second  problem of PIMC is non-locality. It happens that a move which is optimal at a node is not the best move to backup from a global perspective. In some cases it is better to keep a locally suboptimal move that gives a better result at the root of the search tree than the locally optimal one. $\alpha\mu$ addresses this problem by backing up Pareto fronts instead of vectors.

\subsection{Previous Work: the $\alpha\mu$ Algorithm}

The $\alpha\mu$ algorithm repairs the strategy fusion and non-locality problems of PIMC by searching multiple moves ahead and by manipulating Pareto fronts at Max and Min nodes. The $\alpha\mu$ algorithm assumes that the defence has perfect information whereas the declarer has incomplete information.

At Max nodes, each possible move returns a Pareto front. The final Pareto front for a node is the union of all the Pareto fronts returned by the search beginning with each of the different possible moves. The idea is to keep all the possible options for Max, i.e. Max has the choice between all the vectors of the overall Pareto front. In order to optimize computations and memory, vectors that are dominated by another vector in the same Pareto front are removed.

The Min players can choose different moves in different possible worlds, so they take the minimum outcome over all possible moves for a possible world. When they can choose between two vectors they take for each index the minimum between the two values at this index of the two vectors.

The $\alpha\mu$ algorithm is related to GIB single dummy solver \cite{ginsberg2001}. However the single dummy solver of Ginsberg is limited to the endgame and its complexity explodes rapidly for earlier states that $\alpha\mu$ handles easily at the price of incompleteness. Given enough time and using all possible worlds, $\alpha\mu$ converges to the single dummy solver results.

\subsection{Generation of Possible Worlds}

The possible worlds are generated using random generation followed by verification of the constraints. We currently use three types of constraints: the constraints coming from the bidding phase, the constraints on the West hand due to following rules for the opening lead, the constraints due to sluff.

\subsection{Early Cut}

If a Pareto front at a Min node is dominated by the Pareto front of the upper Max node it can safely be cut since the evaluation is optimistic for the Max player. A deeper search will always return a worse (or the same) result for the Max player due to strategy fusion.

\begin{figure}
  \centering
  \caption{Example of an early cut at node c.}
  \label{figureEarlyCut}
\begin{tikzpicture}[level/.style={sibling distance=40mm/#1}]
\node [rectangle,draw,dashed] (z) {a}
  child {node [circle,draw,label=right:{$\{[1~1~0],[0~1~1]\}$}] (b) {b}
     child {node (d) {}} 
     child {node (e) {}} 
  }
  child {node [circle,draw,label=right:{$[1~1~0] \rightarrow cut$}] (c) {c}
  };
\end{tikzpicture}
\end{figure}
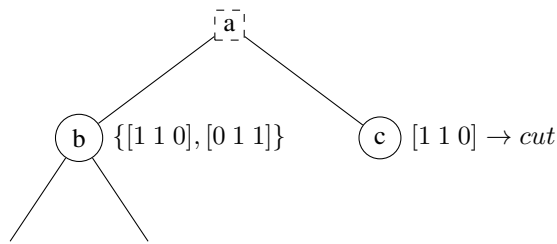

Figure \ref{figureEarlyCut} gives an example of an early cut at a Min node. The root node $a$ is a Max node, the first move played at $a$ returned $\{[1~1~0],[0~1~1]\}$. The second move is then tried leading to node $c$ and the initial Pareto front calculated with double dummy searches at node $c$ is [1 1 0]. It is dominated by the Pareto front of node $a$ so node $c$ can be cut.

\subsection{Root Cut}

If a move at the root of $\alpha\mu$ for $M$ Max moves gives the same probability of winning than the best move of the previous iteration of iterative deepening for $M-1$ Max moves, the search can be safely be stopped since it is not possible to find a better move.

\section{Optimizations}

In this section we present different improvements of the $\alpha\mu$ algorithm that make it faster but do not change the result of a search.

\subsection{Maintaining Useful Worlds}

As we descend the tree it is possible to remove the worlds that are useless to evaluate, allowing us to run fewer DDS evaluations at the leaves, and enabling cuts.

A transposition table perfectly recalls the results of the previous search at nodes. For example if the DDS result stored in the transposition table for a world at a Min node is zero, it will also be zero in the whole subtree. It is therefore useless to calculate it again and the world can be marked as useless.

If at a Min node the maximum value of a world for the current Pareto front is zero, the world can be marked as useless as it will always have a zero value in the Pareto front returned by the node.

At the leaves only useful worlds are solved by DDS.

At Min nodes it is useless to unify the set of possible moves with the possible moves of the useless worlds. We thus only take the union of the possible moves in useful worlds.



\subsection{World Cuts}

Cuts due to having only zero or one useful world are called world cuts.

If the search is at a node without useful worlds it can be safely cut. 

If there are no useful worlds left, search can be cut as the Pareto front is known to be the vector with only zeros.

Search can also be cut if there is only one useful world left. The reason for this is that all of the useless worlds will eventually evaluate to zero at the root, and so we only need to compute the DDS result associated to the single useful world and we can return a single vector containing the result for the useful world.

Figure \ref{figure0WorldCut} gives an example of a world cut with no useful worlds and figure \ref{figure1WorldCut} gives an example of a world cut with one useful world.

\begin{figure}
  \centering
  \caption{Example of a world cut with no useful worlds at node c.}
  \label{figure0WorldCut}
\begin{tikzpicture}[level/.style={sibling distance=60mm/#1}]
\node (root) {...}
child {
  node [circle,draw,label=right:{$\{[1~0~0]\}$}] (z) {a}
  child {node [rectangle,draw,dashed,label=right:{$\{[1~0~0]\}$}] (b) {b}
    child {node (b1) {$[1~0~0]$}}
    }
  child {node [rectangle,draw,dashed,label=right:{$[$x$~-~-] \rightarrow [$0$~$0$~0]$}] (c) {c}
    child {node (c1) {$cut$}}
  }};
\end{tikzpicture}
\end{figure}
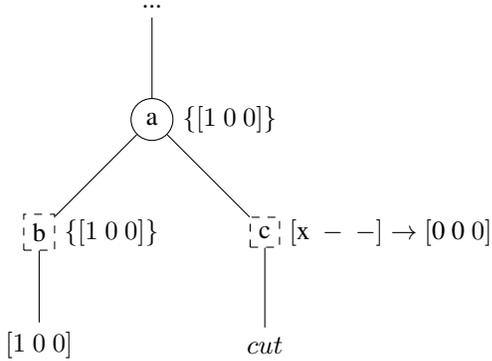

\begin{figure}
  \centering
  \caption{Example of a world cut with one useful world at node c.}
  \label{figure1WorldCut}
\begin{tikzpicture}[level/.style={sibling distance=60mm/#1}]
\node (root) {...}
child {
  node [circle,draw,label=right:{$\{[1~0~1],[0~1~1]\}$}] (z) {a}
  child {node [rectangle,draw,dashed,label=right:{$\{[1~0~1],[0~1~1]\}$}] (b) {b}
    child {node (b1) {$[1~0~1]$}}
    child {node (b2) {$[0~1~1]$}}
    }
  child {node [rectangle,draw,dashed,label=right:{$[$x$~$x$~?] \rightarrow [$x$~$x$~DDS]$}] (c) {c}
    child {node (c1) {$cut$}}
  }};
\end{tikzpicture}
\end{figure}
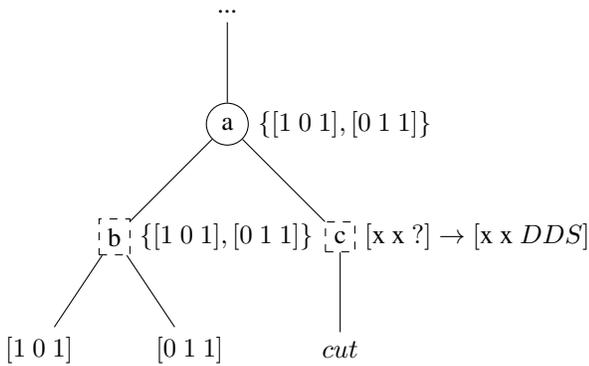







\subsection{Calculating Vectors at Interior Nodes}

It happens that some moves are not explored due to root cut for example. However when the evaluation of the best move at the root gets worse with more search, these moves are then explored to a depth greater than one. In order to verify if an early cut is possible at a depth smaller than the search depth for these moves, the algorithm has to calculate the Pareto fronts at smaller depths. If it can cut with the resulting front it has gained much search efficiency.

We call this optimization the \textit{Empty Entry} optimization.

\subsection{Cut on Win}

If at a Max node a move is found for which all worlds are won for the current search depth, the search can be cut, as more search cannot improve on this result.


\subsection{$\alpha$ Cut}

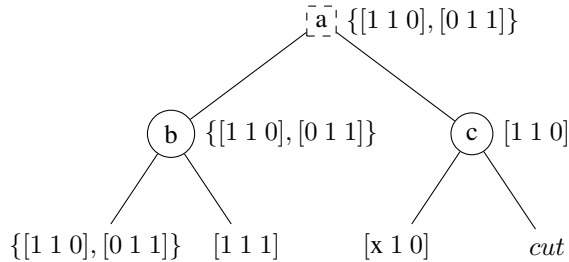
\begin{figure}
  \centering
  \caption{Example of an $\alpha$ cut at node c.}
  \label{figureAlphaCut}
\begin{tikzpicture}[level/.style={sibling distance=40mm/#1}]
\node [rectangle,draw,dashed,label=right:{$\{[1~1~0],[0~1~1]\}$}] (z) {a}
  child {node [circle,draw,label=right:{$\{[1~1~0],[0~1~1]\}$}] (b) {b} 
     child {node (d) {$\{[1~1~0],[0~1~1]\}$}}
     child {node (e) {$[1~1~1]$}}
  }
  child {node [circle,draw,label=right:{$[1~1~0]$}] (c) {c}
      child {node (f) {$[$x$~1~0]$ }}
      child {node (g) {$cut$}}
  };
\end{tikzpicture}
\end{figure}

Figure \ref{figureAlphaCut} gives an example of an $\alpha$ cut at a Min node. The root node a is a Max node, the first move played at $a$ returned $\{[1~1~0],[0~1~1]\}$. The second move is then tried leading to node $c$ and the initial Pareto front calculated with double dummy searches at node $c$ is [1 1 1]. The first Min move at $c$ returns the [x 1 0] Pareto front. The x means that the Min move is not possible in the first world and thus there is no evaluation associated to this world for the first Min move. The current Pareto front at node $c$ is
updated by taking the minimum in all the possible and evaluated worlds of the outcomes, leading to a [1 1 0] Pareto front. This Pareto front is dominated by the Pareto front of the left move at node $a$ and therefore the search can be cut.

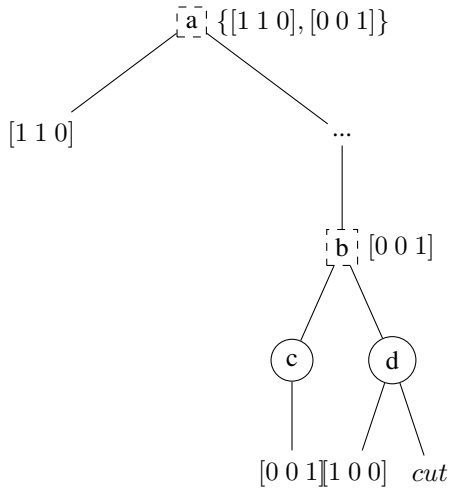
\begin{figure}
  \centering
  \caption{Example of a deep $\alpha$ cut at node d.}
  \label{figureDeepAlphaCut}
\begin{tikzpicture}[level/.style={sibling distance=40mm/#1}]
\node [rectangle,draw,dashed,label=right:{$\{[1~1~0],[0~0~1]\}$}] (a) {a}
  child {node (z) {$[1~1~0]$}}
  child {node (y) {...}
      child {node [rectangle,draw,dashed,label=right:{$[0~0~1]$}] (b) {b}
        child {node [circle,draw] {c} child {node (cc) {$[0~0~1]$}}}
        child {node [circle,draw] {d} 
            child {node (d1) {$[1~0~0]$}}
            child {node (d2) {$cut$}}}
        }
  };
\end{tikzpicture}
\end{figure}

Figure \ref{figureDeepAlphaCut} gives an example of a deep $\alpha$ cut at a Min node. For each max node earlier in the path, if there exists one node whose Pareto front dominates currently evaluated front, further search can be avoided.

Algorithm \ref{AlphaCut} shows how to go up in the tree from a candidate node to check if the candidate node can be cut. The principle of the algorithm is to check all upper Max nodes to see if one of them dominates the current candidate node. We first save the current node to be evaluated against upper node (line 2), we then crawl the tree upward until reaching a max node (line 4 to 7). If we do not find any Max node it means that we reached the tree root and can return false, otherwise we test for Pareto dominance (line 11). If the upper max node does not dominate the candidate node, we search for the next Max node in the tree (line 3).



\begin{algorithm}
\begin{algorithmic}[1]
\STATE{$\alpha$cut(node)}
\begin{ALC@g}
\STATE{candidate $\leftarrow$ node}
\WHILE{node exist}
\STATE{node $\leftarrow$ parent(node)}
\WHILE{node exist \AND Min parent(node)}
\STATE{node $\leftarrow$ parent(node)}
\ENDWHILE
\IF{node doesn't exist}
\RETURN false
\ENDIF
\IF{candidate.front $\leq$ node.front}
\RETURN true
\ENDIF
\ENDWHILE
\RETURN false
\end{ALC@g}
\end{algorithmic}
\caption{\label{AlphaCut}The deep $\alpha$ cut}
\end{algorithm}

\subsection{The $\alpha\mu$ Algorithm with Cuts}

Algorithm \ref{AlphaMuTT} gives the $\alpha\mu$ algorithm with cuts. $M$ is the number of Max moves to search. If this number is equal to zero or if the state is terminal or if there is at most one useful world left the search is stopped (lines 2-5).

At a Min node if there is an early cut the search stops (lines 9-11). Otherwise the union of the sets of legal moves of the useful worlds is calculated (lines 12-16). All moves are tried, maintaining the possible worlds (line 21), making a recursive call (line 22), updating the Pareto front (line 23), updating the useful worlds (line 24) and making a cut if possible (lines 25-27).

At a Max node similar operations are performed, the root cut is also tested (lines 43-47).

\begin{algorithm}
\begin{algorithmic}[1]
\STATE{Function $\alpha\mu$ ($state,M,Worlds,\alpha$)}
\begin{ALC@g}
\IF{$stop (state,M,Worlds,result)$}
\STATE{update the transposition table}
\RETURN $result$
\ENDIF
\STATE{$t \leftarrow$ entry in the transposition table}
\IF{Min node}
\STATE{$mini \leftarrow \emptyset$}
\IF{$t.front \leq \alpha$}
\RETURN $mini$
\ENDIF
\STATE{$Worlds$ = updateUsefulWorlds($t.front,Worlds$)}
\STATE{$allMoves \leftarrow \emptyset$}
\FOR{$w \in Worlds$}
\STATE{$l \leftarrow$ legalMoves ($w$)}
\STATE{$allMoves = allMoves \cup l$}
\ENDFOR
\STATE{move $t.move$ in front of $allMoves$}
\FOR{$move \in allMoves$}
\STATE{$s \leftarrow$ play ($move,state$)}
\STATE{$W_1 \leftarrow \{w \in Worlds : move \in w\}$}
\STATE{$f \leftarrow \alpha\mu$ ($s,M,W_1,\emptyset$)}
\STATE{$mini \leftarrow$ min($mini,f$)}
\STATE{$Worlds$ = updateUsefulWorlds($mini,Worlds$)}
\IF{$mini \leq$ to an upper front}
\STATE{break}
\ENDIF
\ENDFOR
\STATE{update the transposition table}
\RETURN $mini$
\ELSE
\STATE{$front \leftarrow \emptyset$}
\FOR{$w \in Worlds$}
\STATE{$l \leftarrow$ legalMoves ($w$)}
\STATE{$allMoves = allMoves \cup l$}
\ENDFOR
\STATE{move $t.move$ in front of $allMoves$}
\FOR{$move \in allMoves$}
\STATE{$s \leftarrow$ play ($move,state$)}
\STATE{$W_1 \leftarrow \{w \in Worlds : move \in w\}$}
\STATE{$f \leftarrow \alpha\mu$ ($s,M-1,W_1,front$)}
\STATE{$front \leftarrow$ max($front,f$)}
\IF{root node}
\IF{$\mu(front) = \mu$ of previous search}
\STATE{break}
\ENDIF
\ENDIF
\ENDFOR
\STATE{update the transposition table}
\RETURN $front$
\ENDIF
\end{ALC@g}
\end{algorithmic}
\caption{\label{AlphaMuTT}The $\alpha\mu$ search algorithm with cuts.}
\end{algorithm}

\section{Experimental Results}

In the experiments we have $\alpha\mu$ play either the 3NT contract or the 7NT contract against PIMC with 20 simulated worlds at each decision point, or against WBridge5. 

\subsection{Search Times with and without Optimizations}




We fix a number of cards for the initial state and we play a game from this state using $\alpha\mu$, recording the average time per move. The initial number of cards is either 32 or 52. We play 100 games, and thus 3200 searches for 32 cards and 5200 searches for 52 cards. We make experiments both for 20 and 40 simulated worlds at each decision point. In all experiments the early cut and the root cut are enabled.

Table \ref{tableTime-3-20-52} gives the average times used by the program with and without optimizations for deals with 52 cards, 20 worlds and three Max moves. Cards is the number of Cards for the starting hand of each game. M is the number of Max moves allowed during the search. Worlds is the number of worlds used by the search. U is for maintaining useful worlds. E is the use of the Empty Entry optimization. $\alpha$ is the use of the $\alpha$ cut. W is the use of the World cuts. Win is the use of Cut on Win.

Table \ref{tableTime-3-20-32} gives the times for deals with 32 cards, 20 worlds and three Max moves.

Table \ref{tableTime-3-40-32} gives the times for deals with 32 cards, 40 worlds and three Max moves. We see that the speedup factor carries through to the case with twice as many simulated worlds.





\begin{table}
  \centering
  \caption{Comparison of the average time per move of different
  configurations of $\alpha\mu$ with three Max moves and 20 worlds on deals with 52 cards.}
  \label{tableTime-3-20-52}
  \begin{tabular}{rrrrrrrrrrr}
Cards & M & Worlds & U & E & $\alpha$ & W & Win &  Time\\
      &   &        &    &   &         &   &     &      \\
   52 & 3 &     20 &  n & n &       n & n &   n & 3.020\\
   52 & 3 &     20 &  y & n &       n & n &   n & 2.916\\
   52 & 3 &     20 &  n & y &       n & n &   n & 3.321\\
   52 & 3 &     20 &  n & n &       y & n &   n & 3.342\\
   52 & 3 &     20 &  n & n &       n & y &   n & 3.648\\
   52 & 3 &     20 &  n & n &       n & n &   y & 2.837\\
   52 & 3 &     20 &  y & y &       y & y &   y & \bf 1.032\\
   52 & 3 &     20 &  n & y &       y & y &   y & 3.230\\
   52 & 3 &     20 &  y & n &       y & y &   y & 1.438\\
   52 & 3 &     20 &  y & y &       n & y &   y & 1.469\\
   52 & 3 &     20 &  y & y &       y & n &   y & 1.365\\
   52 & 3 &     20 &  y & y &       y & y &   n & 3.566\\
  \end{tabular}
\end{table}

\begin{table}
  \centering
  \caption{Comparison of the average time per move of different
  configurations of $\alpha\mu$ with three Max moves and 20 worlds on deals with 32 cards.}
  \label{tableTime-3-20-32}
  \begin{tabular}{rrrrrrrrrrr}
Cards & M & Worlds & U & E & $\alpha$ & W & Win &  Time\\
      &   &        &    &   &         &   &     &      \\
   32 & 3 &     20 &  n & n &       n & n &   n & 2.016\\
   32 & 3 &     20 &  y & n &       n & n &   n & 1.465\\
   32 & 3 &     20 &  n & y &       n & n &   n & 1.739\\
   32 & 3 &     20 &  n & n &       y & n &   n & 1.704\\
   32 & 3 &     20 &  n & n &       n & y &   n & 1.972\\
   32 & 3 &     20 &  n & n &       n & n &   y & 1.259\\
   32 & 3 &     20 &  y & y &       y & y &   y & \bf 0.605\\
   32 & 3 &     20 &  n & y &       y & y &   y & 0.989\\
   32 & 3 &     20 &  y & n &       y & y &   y & 0.648\\
   32 & 3 &     20 &  y & y &       n & y &   y & 0.647\\
   32 & 3 &     20 &  y & y &       y & n &   y & 0.629\\
   32 & 3 &     20 &  y & y &       y & y &   n & 1.207\\
  \end{tabular}
\end{table}

\begin{table}
  \centering
  \caption{Comparison of the average time per move of different
  configurations of $\alpha\mu$ with three Max moves and 40 worlds on deals with 32 cards.}
  \label{tableTime-3-40-32}
  \begin{tabular}{rrrrrrrrrrr}
Cards & M & Worlds & U & E & $\alpha$ & W & Win &  Time\\
      &   &        &    &   &         &   &     &      \\
   32 & 3 &     40 &  n & n &       n & n &   n & 4.381\\
   32 & 3 &     40 &  y & n &       n & n &   n & 2.891\\
   32 & 3 &     40 &  n & y &       n & n &   n & 3.939\\
   32 & 3 &     40 &  n & n &       y & n &   n & 3.852\\
   32 & 3 &     40 &  n & n &       n & y &   n & 4.338\\
   32 & 3 &     40 &  n & n &       n & n &   y & 2.989\\
   32 & 3 &     40 &  y & y &       y & y &   y & \bf 1.297\\
   32 & 3 &     40 &  n & y &       y & y &   y & 2.420\\
   32 & 3 &     40 &  y & n &       y & y &   y & 1.382\\
   32 & 3 &     40 &  y & y &       n & y &   y & 1.410\\
   32 & 3 &     40 &  y & y &       y & n &   y & 1.329\\
   32 & 3 &     40 &  y & y &       y & y &   n & 2.443\\
  \end{tabular}
\end{table}

\subsection{Games Against WBridge 5}

\begin{table}
  \centering
  \caption{Results of duplicate games against WBridge5 and PIMC on 10 000 deals at 7 No Trump.}
  \label{tableWB5}
  \begin{tabular}{rrrrrrrrrrr}
 P1          &   P2 &    D & $\neq$ & Winrate & $\sigma$\\
             &      &      &       &         &         \\
 $\alpha\mu$ &  WB5 & PIMC &   812 &   0.567 &  0.0174 \\
 $\alpha\mu$ &  WB5 &  WB5 &   755 &   0.428 &  0.0180 \\
 $\alpha\mu$ &  WB5 &  DDS &   567 &   0.697 &  0.0193 \\
 $\alpha\mu$ & PIMC & PIMC &   757 &   0.551 &  0.0181 \\
 $\alpha\mu$ & PIMC &  WB5 &   687 &   0.569 &  0.0189 \\
 $\alpha\mu$ & PIMC &  DDS &   467 &   0.647&   0.0221 \\
  \end{tabular}
\end{table}

\begin{figure}
\centering
\includegraphics[scale=0.5]{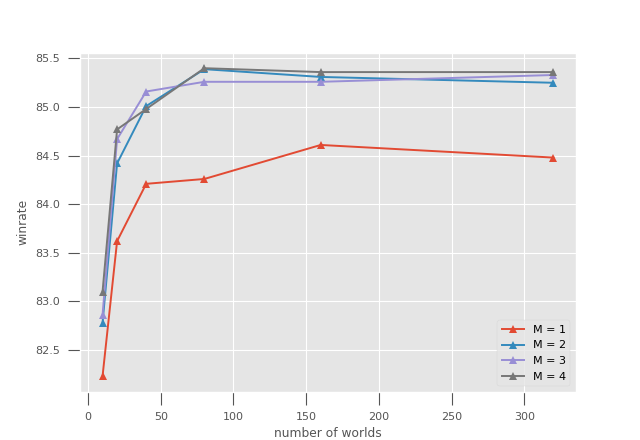}
\caption{The evolution of the winrate with the number of worlds and the depth of the search.}
\label{winrate}
\end{figure}

\begin{figure}
\centering
\includegraphics[scale=0.5]{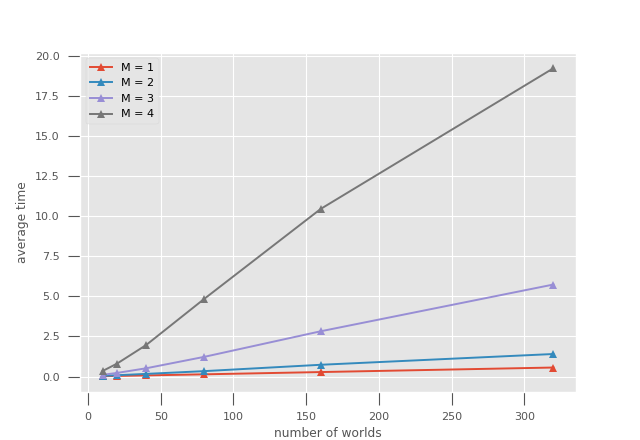}
\caption{The evolution of the thinking time with the number of worlds and the depth of the search.}
\label{time}
\end{figure}

In order to compare two algorithms we use duplicate games. It means that the two algorithms play the same 10 000 deals against the same algorithm for defense. Table \ref{tableWB5} shows the results of duplicate games against different algorithms. The P1 column shows the Max player associated to the winrate. The P2 column shows the other player to which it is compared. The D column indicates the algorithm used for playing the defense. P1 and P2 are playing the declarer for a 7 No Trump contract.  The $\neq$ column gives the number of games where P1 has a different outcome than P2. The Winrate column shows the percentage of these different games that were won by P1 and lost by P2. The standard deviation is given in column $\sigma$. The players are $\alpha\mu$, WBridge5 (WB5) the winner of the 2016, 2017 and 2018 World Computer-Bridge Championships, Perfect Information Monte Carlo (PIMC), and the Double Dummy Solver (DDS) which has complete information and plays knowing the hands of all players.

We can see on the first line that $\alpha\mu$ outperforms WBridge 5 against PIMC as the defense, but we see on the second line that WBridge5 performs better against itself playing as the defense. After discussing with the author of WBridge5, this may be due to the fact that WBridge5 as a declarer partly models itself as the defense during its search. The third line shows that $\alpha\mu$ performs much better than WBridge5 against a perfect complete information defense (DDS). The next three lines deal with PIMC as the second player of the duplicate games. $\alpha\mu$ performs better than PIMC against the three defenders, and especially against DDS (as in the duplicate with WBridge5).

Figure \ref{winrate} shows the evolution of the winning percentage of different versions of $\alpha\mu$ against WBridge5. The x-axis is the number of worlds used by $\alpha\mu$. It starts at 20 worlds and doubles the number of worlds until 320 worlds. There are curves for each number of Max moves between 1 and 4. Note that $\alpha\mu$ with 1 max Move is PIMC. We can observe that the curves are asymptotic and that the asymptote of PIMC is below the asymptote of $\alpha\mu$ with numbers of Max moves greater than 1. Going from 2 Max moves to 3 or 4 does not improve the results.

Figure \ref{time} gives the average time per move of the different versions of $\alpha\mu$. We can observe that $\alpha\mu$ with two Max moves is close to PIMC. However for equivalent thinking times for $\alpha\mu$ with two Max moves and for PIMC with 320 worlds, if we refer to figure \ref{winrate}, $\alpha\mu$ with two Max moves has better results than PIMC for equivalent times.

\subsection{Leaf Parallelization}

Previous experiments did not use parallelization. We now present a simple parallelization of the algorithms with cuts.

In Monte Carlo Tree Search there are three kinds of parallelization: root, leaf and tree parallelization \cite{cazenave2007parallelization,chaslot2008parallel,CazenaveJ08}. Root parallelization runs independent searches in parallel and sums the result to choose the most simulated move. Tree parallelization has multiple threads sharing a common tree. Leaf parallelization parallelizes the playouts at the leaves of the tree.

In order to speedup $\alpha\mu$, we are interested in the simplest form of parallelization, namely leaf parallelization. When reaching a leaf, $\alpha\mu$ does an $\alpha\beta$ complete information double dummy search on each possible world compatible with the cards played so far. These $\alpha\beta$ searches are independent and can safely be run in parallel. This is what we call the leaf parallelization of $\alpha\mu$.

In order to optimize Leaf Parallelization each thread is assigned the next available world as soon as the thread becomes available. There is a mutual exclusion on the assignment of a world to a thread.

Table \ref{tableLeaf} gives the comparison of the times spent per move for $\alpha\mu$ with and without parallelization for different numbers of Max moves and different depth. Leaf parallelization was done with OpenMP and mutual exclusion to choose the next available thread. We can see that there is little to gain for PIMC with 20 worlds, going from 0.019 seconds for 1 thread to 0.015 seconds for 6 threads. For PIMC with 80 worlds going from 1 to 6 threads halves the average time per move. Similar speedups occur for $\alpha\mu$ with two Max moves: the speedup factor is approximately two for 80 worlds. Leaf parallelized $\alpha\mu$ with two Max moves and 40 worlds has approximately the same thinking time as PIMC with 160 worlds, and we can see in figure \ref{winrate} that it has a better winrate.

\begin{table}
  \centering
  \caption{Comparison of the average time per move with Leaf Parallelization using 1 thread and 6 threads.}
  \label{tableLeaf}
  \begin{tabular}{llllllllll}
Cards & M & Worlds & 1 thread &  6 threads \\
      &   &        &    &    &      \\
   52 & 1 &     20 &  0.019 & 0.015 \\
   52 & 1 &     40 &  0.041 & 0.024 \\
   52 & 1 &     80 &  0.083 & 0.041 \\
   52 & 1 &    160 &  0.160 & 0.078 \\
   52 & 2 &     20 &  0.067 & 0.042 \\
   52 & 2 &     40 &  0.145 & 0.081 \\
   52 & 2 &     80 &  0.333 & 0.152 \\
  \end{tabular}
\end{table}

\section{Conclusion}

We presented five different optimizations that speed up search for the $\alpha\mu$ algorithm, and showed how each optimization contributes to the final speedup when combined. In our experiments, the algorithm with the optimizations was shown to be three times faster than without the optimizations. The optimized algorithm has also been parallelized with leaf parallelization which gave a speedup factor of 2. The evolution of the winrate and the thinking time with the number of worlds has shown an asymptotic behavior both for PIMC and $\alpha\mu$ with different numbers of Max moves. The asymptote of $\alpha\mu$ is higher than the asymptote of PIMC, meaning that for usual thinking times and number of worlds, optimized $\alpha\mu$ performs better than PIMC. Moreover, optimized $\alpha\mu$ outperforms the former computer Bridge world champion WBridge5 in the 7NT contract when the defense is played by PIMC or DDS.

\section*{Acknowledgment}

Thanks to Dan Braun for proofreading.


\bibliography{main}

\end{document}